\documentclass[journal]{IEEEtran}
\usepackage{amsmath,amsfonts}
\usepackage{algorithmic}
\usepackage{algorithm}
\usepackage{array}
\usepackage{textcomp}
\usepackage{stfloats}
\usepackage{url}
\usepackage{verbatim}
\usepackage{graphicx}
\usepackage{cite}
\usepackage{multirow}
\usepackage{balance}
\usepackage{threeparttable}
\usepackage{bbm}
\usepackage{bm}
\usepackage{booktabs}
\usepackage{diagbox}
\usepackage[table]{xcolor}
\begin{document}

\title{A Multi-level Supervised Contrastive Learning Framework for Low-Resource Natural Language Inference}

\author{Shu'ang Li, Xuming Hu, Li Lin, Aiwei Liu, Lijie Wen\thanks{
The work was supported by the National Key Research and Development Program of China (No. 2019YFB1704003), the National Nature Science Foundation of China (No. 71690231 and No. 62021002), NSF under grants III-1763325, III-1909323, III-2106758, SaTC-1930941, Tsinghua BNRist and Beijing Key Laboratory of Industrial Bigdata System and Application. (Corresponding authors: Lijie Wen.)

Shu'ang Li, Xuming Hu, Li Lin, Aiwei Liu and Lijie Wen are with School of Software, Tsinghua University, Beijing, China (e-mail:\{lisa18, hxm19, lin-l16, liuaw20\}@mails.tsinghua.edu.cn; wenlj@tsinghua.edu.cn).}, Philip S. Yu, \IEEEmembership{Life Fellow, IEEE}\thanks{
Philip S. Yu is with the Department of Computer Science, University of Illinois at Chicago, Chicago, IL 60607 USA (e-mail: psyu@uic.edu).

This work has been submitted to the IEEE for possible publication. Copyright may be transferred without notice, after which this version may no longer be accessible.}
}



\maketitle

\begin{abstract}
Natural Language Inference (NLI) is a growingly essential task in natural language understanding, which requires inferring the relationship between the sentence pairs (\textbf{premise} and \textbf{hypothesis}). Recently, low-resource natural language inference has gained increasing attention, due to significant savings in manual annotation costs and a better fit with real-world scenarios. Existing works fail to characterize discriminative representations between different classes with limited training data, which may cause faults in label prediction. 
Here we propose a multi-level supervised contrastive learning framework named MultiSCL for low-resource natural language inference. MultiSCL leverages a sentence-level and pair-level contrastive learning objective to discriminate between different classes of sentence pairs by bringing those in one class together and pushing away those in different classes.  
MultiSCL adopts a data augmentation module that generates different views for input samples to better learn the latent representation. The pair-level representation is obtained from a cross attention module. We conduct extensive experiments on two public NLI datasets in low-resource settings, and the accuracy of MultiSCL exceeds other models by 3.1\% on average. Moreover, our method outperforms the previous state-of-the-art method on cross-domain tasks of text classification.

\end{abstract}


\section{Introduction}
\label{sec:intro}
\IEEEPARstart{N}{atural} Language Inference (NLI), also known as Recognizing Textual Entailment (RTE), is a key topic in the research field of natural language understanding~\cite{maccartney-manning-2008-modeling,shen2021towards}, and could support tasks such as question answering, reading comprehension, document summarization and relation extraction~\cite{dagan2013recognizing, hu2020selfore, hu2021gradient, hu-etal-2021-semi-supervised}. In NLI scenarios, the model is given a pair of sentences, namely \textbf{premise} and \textbf{hypothesis}, and asked to infer the relationship between them from a set of relationships, including \texttt{entailment}, \texttt{contradiction} and \texttt{neutral}. Several concrete examples are illustrated in Table \ref{table1}.

\begin{table}[!htp]
\centering
\caption{Examples of the Relationship between \textbf{Premise} and \textbf{Hypothesis}: \underline{\textbf{E}}ntailment, \underline{\textbf{C}}ontradiction, and \underline{\textbf{N}}eutral}\label{table1}
\resizebox{0.9\linewidth}{!}{%
\begin{tabular}{lll}
\toprule
\textbf{Premise} & Two men on bicycles competing in a race. &   \\ \midrule
\multirow{3}{*}{\textbf{Hypothesis}}  & People are riding bikes.                 & E \\
                                      & Men are riding bicycles on the streets.  & C \\
                                      & A few people are catching fish.           & N \\ \bottomrule
\end{tabular}%
}
\end{table}

Large annotated datasets, such as SNLI~\cite{bowman-etal-2015-large} and MultiNLI datasets~\cite{williams2018broad} have been available in recent years, making it possible to apply sophisticated deep learning models.
These neural network models require a large number of training parameters to achieve good results in NLI~\cite{chen2018neural,gong2018natural}. However, large-scale datasets are obtained from a large number of manual annotations and have a high annotation cost. Therefore, natural language inference for low-resource scenarios has gained more widespread attention in recent years. Compared with traditional task scenarios, NLI in low-resource scenarios focuses on using a small amount of manually annotated data to achieve similar results as the full amount of data. This can save a large number of manual annotation costs and is more in line with realistic application scenarios, thus with high research value and practical application value~\cite{hedderich2021survey}.

Recent work has shown advantages of generative classifiers in term of low-resource and robustness. Ding et al. \cite{ding2020discriminatively} propose a generative classifier that defines the conditional probabilities assumed given the premises and labels and has better performance in very few labeled data settings.  Liu et al.~\cite{liu2019multi} propose a multi-task deep neural network for learning semantic representations across multiple natural language understanding tasks to enhance the semantic representation in low-resource scenarios. The network not only utilizes a large amount of cross-task data but also benefits from regularization effects to learn more general representations that can be adapted to NLI in fewer sample scenarios. However, these methods only use the feature of the sentence pair itself to predict the class, without considering the comparison between the sentence pairs in different classes. They fail to characterize discriminative representations between different classes with limited training data, which may cause faults in label prediction. 

Many recent works explored using contrastive learning to tackle this problem. Contrastive learning is a popular technique in computer vision area \cite{he2020momentum, chen2020simple, khosla2020supervised} and the core idea is to learn a function that maps positive pairs closer together in the embedding space, while pushing apart negative pairs. A contrastive objective is used by \cite{gao2021simcse} to fine-tune pre-trained language models to obtain sentence embeddings with the relationship of sentences in NLI. The model achieved state-of-the-art performance in sentence similarity tasks. Yan et al. \cite{yan2021consert} propose a simple but effective training objective based on contrastive learning. It mitigates the collapse of BERT-derived representations and transfers them to downstream tasks. However, these approaches can't distinguish well between the representation of sentence pairs in different classes. 

In our previous work \cite{li2022pair}, we propose a pair-level supervised contrastive learning approach (PairSCL), which obtains new state-of-the-art performance in NLI. However, this method cannot be adapted to low-resource settings for the reason of limited discriminative ability of sentence pairs in very few sample scenarios. Therefore, in this paper, we comprehensively investigate the potential of contrastive learning in low-resource NLI. Based on our analysis of PairSCL that contrastive learning can help discriminate the class of sentence pairs, we propose a multi-level supervised contrastive learning framework named MultiSCL for low-resource NLI. In addition to pair-level contrastive learning, MultiSCL leverages the sentence-level contrastive learning objective to characterize the latent embeddings of sentences in semantic space. Furthermore, to better learn the semantic representation, we adopt a data augmentation module that generates different views for input sentences with sentence-level supervised contrastive learning by regarding the contradiction pairs as negatives, and entailment pairs as positives. The pair-level representation can perceive the class information of sentence pairs and is obtained from the Cross Attention module which captures the relevance and characterizes the relationship between the sentence pair. Then we adopt contrastive learning to differentiate the pair-level representation by capturing the similarity between pairs in one class and contrasting them with pairs in other classes.

For example, the entailment pair ($P_1$, $H_1$) and contradiction pair ($P_2$, $H_2$) are from Table \ref{table1} ($P_1$: Two men on bicycles competing in a race. $H_1$: People are riding bikes. $P_2$: Two men on bicycles competing in a race. $H_2$: Men are riding bicycles on the streets.). For sentence-level contrastive learning, we take advantage of the fact that entailment pairs can be naturally used as positives and the contradiction pairs can be regarded as negatives. We consider $H_1$ as the positive set for $P_1$ and $H_2$ as the negative set for $P_2$. In this way, the encoder can capture the semantic representation of the sentences more accurately. For pair-level contrastive learning, our model regards the pair ($P_2$, $H_2$) as the negative set for the pair ($P_1$, $H_1$) with the representation obtained from Cross Attention module to distinguish the pairs from different classes.

Our contributions can be summarized as follows:
\begin{itemize}
    \item We propose a novel multi-level supervised contrastive learning framework named MultiSCL for low-resource NLI. It applies the sentence-level and pair-level contrastive learning to learn the discriminative representation with limited labeled training data.
    \item We adopt a data augmentation module to generate the different views for input samples. We explore various effective text augmentation strategies for contrastive learning and analyze their effects on low-resource NLI.
    \item We conduct extensive experiments on two public NLI datasets in low-resource settings, and the accuracy of MultiSCL exceeds other models by 3.1\% on average. Moreover, our method outperforms the previous state-of-the-art method on cross-domain tasks of text classification.
\end{itemize}

This paper is substantially an extended version of our previous paper~\cite{li2022pair} that will be published at ICASSP 2022. Compared to the previous version, we make heavy extensions as follows: (1) By adding the sentence-level contrastive learning objective, we propose a new multi-level supervised contrastive learning framework called MultiSCL for low-resource NLI. (2) We adopt a data augmentation module to generate the views for input sentences and explore various effective text augmentation strategies. (3) We conduct extensive experiments on NLI datasets in low-resource scenarios. We conduct experiments on cross-domain datasets to validate the transfer capability of our model. Moreover, We carefully study the components of MutliSCL and provide a detailed ablation study of hyper-parameters.

The structure of this paper is as follows. In Section II, we review the related work to natural language inference and contrastive learning. Section III introduces the architecture of our framework. Section IV presents experimental design details and Section V reports our experimental results and analysis. Finally, in Section VI, we conclude this paper and present some future work.

\section{Related Works}
\subsection{Natural Language Inference}
Early methods for NLI mainly relied on conventional, feature-based methods trained from small-scale datasets~\cite{dagan2013recognizing, marelli2014sick}. The release of large datasets, such as SNLI~\cite{bowman-etal-2015-large} and MultiNLI~\cite{williams2018broad}, made neural network methods feasible. Such methods can be roughly categorized into two classes: sentence embedding bottleneck methods which first encode the two sentences as vectors and then feed them into a classifier for classification~\cite{conneau2017supervised, chen-etal-2017-recurrent-neural, wu2018phrase}, and more general methods which usually involve interactions while encoding the two sentences in the pair~\cite{chen-etal-2017-enhanced, gong2018natural, liang2019asynchronous}. \cite{wang2019improving} enables the use of various kinds of external knowledge bases to retrieve information related to \textbf{premise} and \textbf{hypothesis}. Wang et al. \cite{wang2020knowledge} propose a novel Knowledge Graph-enhanced NLI (KGNLI) model to leverage the usage of background knowledge stored in knowledge graphs in the field of NLI.

Recently, large-scale pre-trained language representation models such as BERT~\cite{devlin2019bert}, GPT~\cite{radford2018improving}, BART~\cite{lewis2020bart}, etc., have achieved dominating performance in NLI. These neural network models have a large number of training parameters to achieve good results in NLI. However, large-scale datasets are obtained from a large number of manual annotations and have a high annotation cost. Therefore, NLI for low-resource scenarios has gained more widespread attention in recent years.

Ding et al. \cite{ding2020discriminatively} propose GenNLI, a generative classifier for NLI tasks. The model defines conditional probabilities assumed given premises and labels, parameterizing the distribution using a sequence-to-sequence model with attention \cite{luong2015effective} and a replication mechanism \cite{gu2016incorporating}. They explore training objectives for discriminative fine-tuning of the generative classifier, comparing several classical discriminative criteria.
Liu et al. \cite{liu2019multi} propose a multi-task deep neural network (MT-DNN) for learning semantic representations across multiple natural language understanding tasks. MT-DNN not only utilizes a large amount of cross-task data but also benefits from regularization effects to learn more general representations that can be adapted to natural language reasoning in very few sample scenarios.
Schick et al. \cite{schick2021exploiting} introduce Pattern Exploiting Training concerning a partially pre-trained language model using "task descriptions" in natural language \cite{radford2019language}. They reformulate a small amount of labeled data into fill-in-the-blank phrases to help the language model understand the given task. 

The above methods only use the feature of the sentence pair itself to predict the class, without considering the comparison between the sentence pairs in different classes. In our work, we propose a multi-level contrastive learning framework named MultiSCL for low-resource NLI. MultiSCL leverages a sentence-level and pair-level contrastive learning objective to learn discriminative representations between different classes.

\subsection{Contrastive Learning}
Contrastive learning has shown promising results in computer vision area in an unsupervised/self-supervised way \cite{he2020momentum, chen2020simple}.
The key idea of contrastive learning is: first create augmentations of original examples, then learn representations by predicting whether two augmented examples are from the same original data example or not.
Dating back to~\cite{hadsell2006dimensionality}, these approaches learn representations by contrasting positive pairs against negative pairs. 
Along these lines, Dosovitskiy et al. \cite{dosovitskiy2015discriminative} propose to treat each instance as a class represented by a feature vector (in a parametric form). Wu et al. \cite{wu2018unsupervised} propose to use a memory bank to store the instance class representation vector, which was an approach adopted and extended in several recent papers~\cite{zhuang2019local, tian2019contrastive}.
He et al. \cite{he2020momentum} propose Momentum Contrast (MoCo) by building a dynamic dictionary with a queue and a moving-averaged encoder and showed state-of-the-art results. Chen et al. \cite{chen2020simple} propose a simple framework for contrastive learning to learn visual representations without specialized architectures or a memory bank.

Some works use contrastive learning to solve Natural Language Processing (NLP) tasks. Qin et al.~\cite{qin2020erica} propose a novel contrastive learning framework in the pre-training phase to obtain a deeper understanding of the entities and their relations in text.
Giorgi et al. \cite{giorgi2020declutr} propose a self-supervised method for minimizing sentence embeddings of textual segments randomly sampled from nearby in the same document and obtained state-of-the-art performance on SentEval~\cite{conneau2018senteval}.
Gunel et al. \cite{gunel2020supervised} propose a supervised contrastive learning (SCL) objective which uses SCL loss combined with cross-entropy loss for the fine-tuning stage. The proposed model shows improved performance on multiple datasets of the GLUE benchmark~\cite{wang2018glue} in both the high-data and low-data regimes.
Yan et al. \cite{yan2021consert} explore a simple but effective sentence-level training objective with various effective text augmentation strategies to generate views for contrastive learning. Suresh et al. \cite{suresh2021not} incorporate inter-class relationships into a supervised contrastive loss by differentiating the weights between different negative samples for fine-grained text classification. Li et al. \cite{li2021learning} adopt supervised contrastive pre-training to capture both implicit and explicit sentiment orientation towards aspects by aligning the representation of implicit sentiment expressions to those with the same label for aspect-based sentiment analysis. Wu et al. \cite{wu2020clear} propose a new framework, combining word-level masked language modeling objectives with sentence-level contrastive learning objective to pre-train a language model. Zhang et al. \cite{zhang2021pairwise} propose an instance discrimination-based approach aiming to bridge semantic entailment and contradiction understanding with high-level categorical concept encoding. Wang et al. \cite{wang2021cline} propose Contrastive Learning with semantIc Negative Examples (CLINE), which constructs semantic negative examples unsupervised to improve the robustness under semantically adversarial attacking. By comparing with similar and opposite semantic examples, the model can effectively perceive the semantic changes caused by small perturbations. Li et al. \cite{li2022pair} propose a pair-level supervised contrastive learning approach. The pair-level representation is obtained by Cross Attention module which can capture the relevance and well characterize the relationship between the sentence pair. 

However, the above methods can't learn effective semantic representations in low-resource scenarios. In our work, we will focus on the use of multi-level contrastive learning for low-resource NLI.
\section{Approach}
\label{sec:model}

\begin{figure*}[ht]
  \centering
  \includegraphics[width=0.95\textwidth]{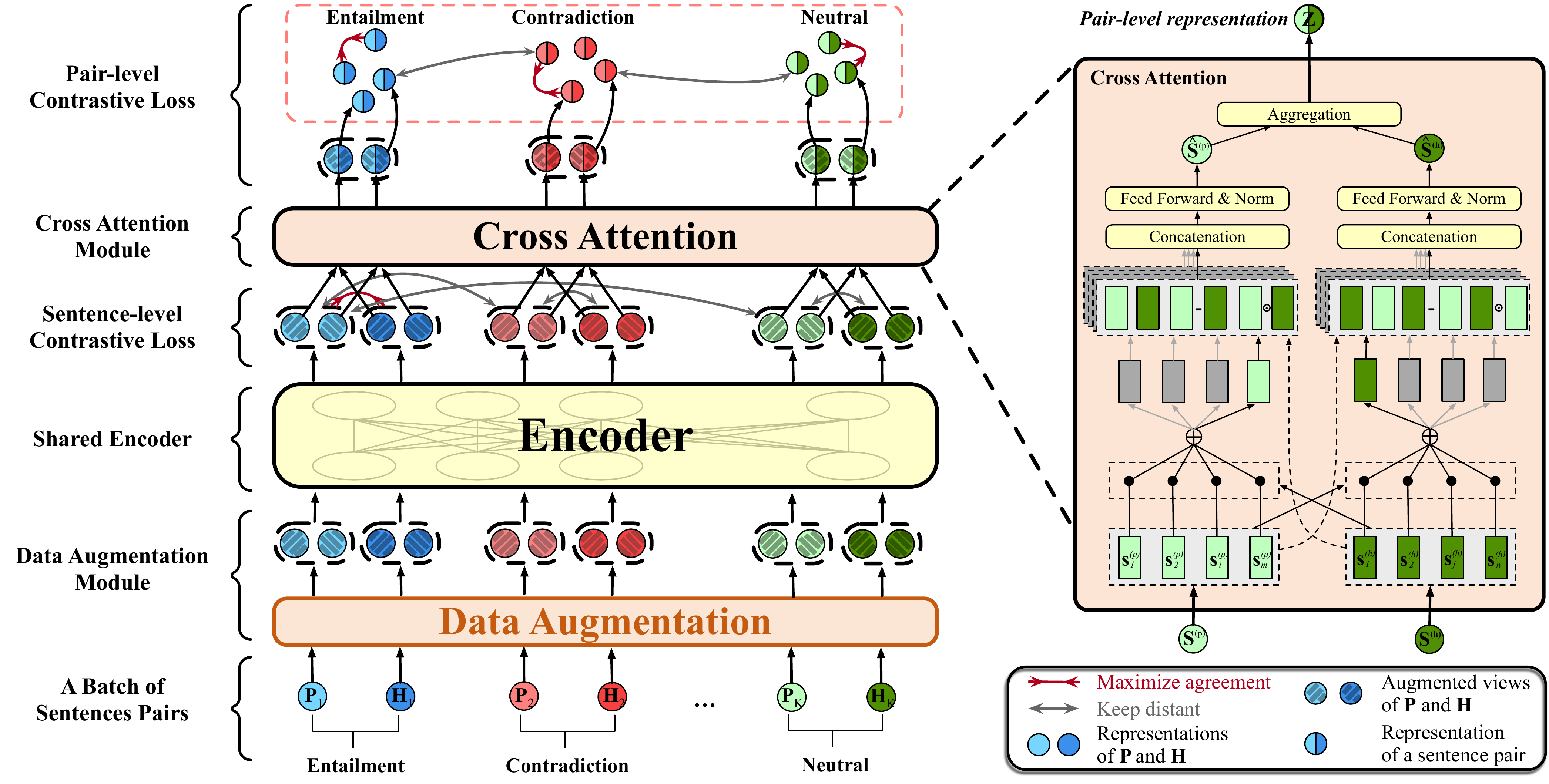} 
  \caption{\label{model} The overall view of MultiSCL. The left part is the main framework of our model. The right part is the detailed structure of Cross Attention module.} 
\end{figure*}

\noindent In this section, we describe our approach MultiSCL. The overall architecture of the model is illustrated in the left part of Figure~\ref{model}. MultiSCL comprises the following four major components: a data augmentation module that generates different views for input samples; an encoder that computes sentence-level representations of \textbf{premise} and \textbf{hypothesis}; a cross attention module to obtain the pair-level representation of the sentence pair and a joint-training layer including the sentence-level and pair-level contrastive learning term and the cross-entropy term.

\subsection{Data Augmentation Module}
\label{data augmentation}
In low-resource NLI scenarios, it's a challenge for a model to learn discriminative representations and infer the relationship between sentences. Therefore, we adopt a data augmentation module to generate different views of sentences to enhance the semantic understanding and inference capability of the model. We explore and test six data augmentation strategies to generate views, including synonym replacement\cite{wei-zou-2019-eda}, reordering\cite{wu2020clear}, word insertion, span deletion, word deletion, dropout\cite{gao2021simcse}, and back translation\cite{edunov2018understanding}.

\textbf{Synonym Replacement} randomly chooses $n$ words from the sentence that are not stop words. Replace each of these words with one of its synonyms chosen at random, so that the augmented data fit the original semantics as closely as possible.

\textbf{Reordering} is another widely-studied augmentation method that can keep the original sentence’s features. We randomly select $n$ pairs of spans and switch them to construct the reordering augmentation in our implementation.

\textbf{Word Insertion} finds the random synonym of $n$ random word in the sentence that is not a stop word. Insert that synonym into a random position in the sentence.

\textbf{Word Deletion} randomly selects $n$ tokens in the sentence and replaces them with a special token  \texttt{[DEL]}, which is similar to the token \texttt{[MASK]} in BERT \cite{devlin2019bert}.

\textbf{Dropout} has been proven an effective augmentation strategy for contrastive learning \cite{gao2021simcse, yan2021consert}. For this setting, we randomly drop elements in the token embedding layer by a specific number $n$ and set their values to zero.

\textbf{Back Translation} first translates the sentence into another language and translates it back to the original language. Then the new sentence is regarded as an augmented sentence of the original.

In our model, we select two augmentation strategies to generate two different views of input texts. We will explore and test the effect of different combinations in our experiment. The number of changed words or tokens for augmentation methods $n$ is based on the length of the sentence $l$ with the formula $n = \eta l$, where $\eta$ is a hyper-parameter that indicates the percent of the changed words in a sentence.

\subsection{Text Encoder}
\label{sec:encoder}
After we get the different views of the sentence, we need to get the context-based semantic information. We give the formal definition of NLI as follows. Each instance in a NLI dataset consists of two sentences and a label indicating the relation between them. Formally, we denote \textbf{premise} as $X^{(p)}=\{x_1^{(p)},x_2^{(p)},\cdots,x_m^{(p)}\}$ and \textbf{hypothesis} as $X^{(h)}=\{x_1^{(h)},x_2^{(h)},\cdots,x_n^{(h)}\}$, where $m$ and $n$ are length of the sentences respectively. The instance in the batch $\mathcal{I}$ is denoted as $(X^{(p)}, X^{(h)}, y)_i$, where $i = \{1, \dots, K\} $ is the indices of the samples and $K$ is the batch-size. After passing the input samples to data augmentation module, we construct the new batch $\tilde{\mathcal{I}}$ with size $2K$ by randomly augmenting twice for all the sentences.
The encoder (e.g., BERT) takes $X^{(p)}, X^{(h)}$ as inputs and computes the semantic representations, denoted as $\mathbf{S}^{(p)} = \{\mathbf{s}_i^{(p)}|\mathbf{s}_i^{(p)} \in \mathbb{R}^{k}, i = 1,2,\cdots, m\}$ and $\mathbf{S}^{(h)} = \{\mathbf{s}_j^{(h)}|\mathbf{s}_j^{(h)} \in \mathbb{R}^{k}, j = 1,2,\cdots, n\}$, where $k$ is the dimension of the encoder’s hidden state.

\subsection{Cross Attention Module}
Different from single sentence classification, we need a proper interaction module to better clarify the sentences pair's relationship for NLI task. In practice, we need to compute token-level weights between words in \textbf{premise} and \textbf{hypothesis} to obtain information about their interaction. Therefore, we introduce Cross Attention module to calculate the co-attention matrix $\mathbf{C} \in \mathbb{R}^{m \times n}$ of the token level. If the value of attention weight is relatively large, the correlation between the words is stronger. Each element $\mathbf{C}_{i,j} \in \mathbb{R}$ indicates the relevance between the i-th word of \textbf{premise} and the j-th word of \textbf{hypothesis}:
\begin{equation}
  \mathbf{C}_{i,j} = \mathbf{P}^Ttanh(\mathbf{W}(\mathbf{s}^{(p)}_i \odot \mathbf{s}^{(h)}_j)),
\end{equation}
where $\mathbf{W} \in \mathbb{R}^{d \times k}$, $\mathbf{P} \in \mathbb{R}^{d}$, and $\odot $ denotes the element-wise production operation. $\mathbf{W}$ and $\mathbf{P}$ are trainable parameters to map the feature in the semantic space, where $k$ is the dimension of the encoder’s hidden state and $d$ is the dimension for nonlinear mapping to a latent representation. Then the attentive matrix could be formalized as:

\begin{align}
  \mathbf{c}_i^{(p)} = softmax(\mathbf{C}_{i,:}),\quad &\mathbf{c}_j^{(h)} = softmax(\mathbf{C}_{:,j}),\\
  \mathbf{s}^{(p)'}_i = \mathbf{S}^{(h)} \cdot \mathbf{c}_i^{(p)},\quad &\mathbf{s}^{(h)'}_i = \mathbf{S}^{(p)} \cdot \mathbf{c}_j^{(h)}.
\end{align}

We further enhance the collected local semantic information:
\begin{gather}
  \mathbf{s}^{(p)''}_i = [\mathbf{s}^{(p)}_i; \mathbf{s}^{(p)'}_i; \mathbf{s}^{(p)}_i - \mathbf{s}^{(p)'}_i; \mathbf{s}^{(p)}_i \odot \mathbf{s}^{(p)'}_i],\\
  \tilde{\mathbf{s}}^{(p)}_i = ReLU(\mathbf{W}^{(p)}_i\mathbf{s}^{(p)''}_i + \mathbf{b}^{(p)}_i),
\end{gather}
where $[\cdot ;\cdot ;\cdot ;\cdot ]$ refers to the concatenation operation. $\mathbf{s}^{(p)}_i-\mathbf{s}^{(p)'}_i$ indicates the difference between the original representation and the \textbf{hypothesis}-information enhanced representation of \textbf{premise}, and $\mathbf{s}^{(p)}_i \odot \mathbf{s}^{(p)'}_i$ represents their semantic similarity. Both values are designed to measure the degree of semantic relevance between the sentence pair. The smaller the difference and the larger the semantic similarity, the sentences pair are more likely to be classified into the Entailment category. The difference and element-wise product are then concatenated with the original vectors ($\mathbf{S}^{(p)}, \mathbf{S}^{(p)^\prime}$). We expect that such operations could help enhance the pair-level information and capture the inference relationships of \textbf{premise} and \textbf{hypothesis}. We get the new representation containing \textbf{hypothesis}-guided inferential information for \textbf{premise}:
\begin{gather}
  \tilde{\mathbf{S}}^{(p)} = (\tilde{\mathbf{s}}^{(p)}_1, \tilde{\mathbf{s}}^{(p)}_2, \dots, \tilde{\mathbf{s}}^{(p)}_m),\\
  \hat{\mathbf{S}}^{(p)} = LayerNorm(\tilde{\mathbf{S}}^{(p)}),
\end{gather}
where $LayerNorm(.)$ is a layer normalization. The result $\hat{\mathbf{S}}^{(p)}$ is a 2D-tensor that has the same shape as $\mathbf{S}^{(p)}$. The representation of \textbf{hypothesis} $\hat{\mathbf{S}}^{(h)}$ is calculated in the same way. 

Then we convert these representations obtained above to a fixed-length vector with pooling. More specifically, we compute max pooling and mean pooling for $\hat{\mathbf{S}}^{(p)}$ and $\hat{\mathbf{S}}^{(h)}$. where $\hat{\mathbf{S}}^{(p)} = \{ \hat{\mathbf{s}}^{(p)}_1, \hat{\mathbf{s}}^{(p)}_2,...,\hat{\mathbf{s}}^{(p)}_m\}$ and $\hat{\mathbf{S}}^{(h)} = \{ \hat{\mathbf{s}}^{(h)}_1, \hat{\mathbf{s}}^{(h)}_2,...,\hat{\mathbf{s}}^{(h)}_n\}$. Formally:
\begin{align}
    \hat{\mathbf{S}}^{(p)}_{\operatorname{mean}}=\sum_{i=1}^{m} \frac{\hat{\mathbf{s}}^{(p)}_i}{m},\quad & \hat{\mathbf{S}}^{(p)}_{\max}=\max_{i=1}^{m} \hat{\mathbf{s}}^{(p)}_i,\\
    \hat{\mathbf{S}}^{(h)}_{\operatorname{mean}}=\sum_{j=1}^{n} \frac{\hat{\mathbf{s}}^{(h)}_j}{n},\quad &
    \hat{\mathbf{S}}^{(h)}_{\max}=\max_{j=1}^{n} \hat{\mathbf{s}}^{(h)}_j.
\end{align}

We aggregate these representations and the pair-level representation $\mathbf{Z}$ for the sentence pair is obtained as follows:
\begin{equation}
  \mathbf{Z} = [\hat{\mathbf{S}}^{(p)}_{\operatorname{mean}}; \hat{\mathbf{S}}^{(p)}_{\max}; \hat{\mathbf{S}}^{(h)}_{\operatorname{mean}}; \hat{\mathbf{S}}^{(h)}_{\max}].
\end{equation}

As described, Cross Attention module can capture the relevance of the sentence pair and well characterize the relationship. Therefore, the pair-level representation can perceive the class information of sentence pairs.

\subsection{Training Objective}

\subsubsection{\textbf{Text-level Supervised Contrastive Loss}}
\label{sec:text-level}
The core idea of contrastive learning is to learn a function that maps positive pairs closer together in the embedding space while pushing apart negative pairs. In general, two variants augmented from the same original sentence form the positive pair, while all other instances from the same batch are regarded as negative samples for them. Especially, the NLI datasets consist of high-quality and crowd-sourced labeled sentence pairs and each can be presented in the form: (\textbf{premise}, \textbf{hypothesis}, label) as described in Section \ref{sec:intro}. Given one premise, human annotators are required to manually write one sentence that is absolutely true (entailment), one that might be true (neutral), and one that is definitely false (contradiction). Thus for each \textbf{premise} and its entailment \textbf{hypothesis}, there is an accompanying contradiction \textbf{hypothesis} and neutral \textbf{hypothesis} (see Table \ref{table1} for an example). Therefore, it is natural to take the entailment \textbf{hypothesis} for \textbf{premise} as its positive set and the contradiction \textbf{hypothesis} as negative set. 

In the training stage, we randomly sample a batch $\mathcal{I}$ of $K$ examples $(X^{(p)}, X^{(h)}, y)_{i \in \mathcal{I} = \{1, \dots, K\}} $ as denoted in Section \ref{sec:encoder}.  After passing the input samples to data augmentation module, we construct the new batch $\tilde{\mathcal{I}}$ with size $2K$ by randomly augmenting twice for all the sentences. We denote the representation obtained from the text encoder of the new batch as $(\mathbf{S}^{(p)}, \mathbf{S}^{(h)})_{i \in \tilde{\mathcal{I}} = \{1, \dots, 2K\}}$. For the premise $\mathbf{S}^{(p)}_i$ in the augmented batch, we denote the set of positives as $\mathbf{S}^{(p)+}_i$, including the augmented views of the entailment \textbf{hypothesis} and the augmented views of the same original \textbf{premise}. The negative set is denoted as $\mathbf{S}^{(p)-}_i$, including the augmented views of the contradiction \textbf{hypothesis}. The sentence-level supervised contrastive loss on the batch $\tilde{\mathcal{I}}$ is defined as:
\begin{equation}
      \ell_{i} = \frac{e^{\operatorname{sim}(\mathbf{S}^{(p)}_i, \mathbf{S}^{(p)+}_i) /\tau}}{\sum_{k=1}^{2K}\mathbbm{1}_{[k \neq i]}(e^{\operatorname{sim}(\mathbf{S}^{(p)}_i, \mathbf{S}^{(p)+}_k)/\tau} + e^{\operatorname{sim}(\mathbf{S}^{(p)}_i, \mathbf{S}^{(p)-}_k)/\tau})},
\end{equation}
where $\operatorname{sim}$(·) indicates the cosine similarity function, $\tau$ controls the temperature.

Finally, we average all $2K$ in-batch losses $\ell_{i}$ to obtain the final sentence-level contrastive loss $\mathcal{L}_{SCL(sent)}$:
\begin{equation}
    \mathcal{L}_{SCL(sent)} = \sum_{i \in \tilde{\mathcal{I}}}-\operatorname{log}  \ell_{i},
\end{equation}

In this way, we can map the representations from the encoder of the semantically similar sentences closer together in the embedding space, while pushing apart irrelevant sentences. Thus, the pair-level representations based on the output of the encoder can better capture the relationships between \textbf{premise} and \textbf{hypothesis}.

\subsubsection{\textbf{Pair-level Supervised Contrastive Loss}}
In \cite{khosla2020supervised}, the authors extended the above loss to a supervised contrastive loss by regarding the samples belonging to the same class as the positive set. Inspired by this, we adopt a supervised contrastive learning objective to align the pair-level representation obtained from Cross Attention module to distinguish sentence pairs from different classes. The pair-level supervised contrastive loss brings the latent representations of pairs belonging to the same class closer together.

In the training stage, we take the augmented batch $\tilde{\mathcal{I}}$ of $2K$ examples $(X^{(p)}, X^{(h)}, y)_{i = \{1, \dots, 2K\}} $ as denoted in Section \ref{sec:text-level}. For the pair $(X^{(p)}, X^{(h)}, y)_i$, we denote the set of positives as $\mathcal{P} = \{p: p\in \mathcal{I}, y_p=y_i\wedge p \neq i\}$, with size $|\mathcal{P} |$. The supervised contrastive loss on the batch $\tilde{\mathcal{I}}$ is defined as:
\begin{align}
  \ell_{i, p} = \frac{e^{\operatorname{sim}(\mathbf{Z}_i \cdot \mathbf{Z}_p/\tau)}}{\sum_{k \in \tilde{\mathcal{I}}/i}e^{\operatorname{sim}(\mathbf{Z}_i \cdot \mathbf{Z}_k/\tau)}},\\
  \mathcal{L}_{SCL(pair)} = \sum_{i \in \tilde{\mathcal{I}}}-\operatorname{log} \frac{1}{|\mathcal{P}|}\sum_{p \in \mathcal{P}} \ell_{i, p},
\end{align}
where $\ell_{i, p}$ indicates the likelihood that pair $i$ is most similar to pair $p$ and $\tau$ is the temperature hyper-parameter. Larger values of $\tau$ scale down the dot-products, creating more difficult comparisons. $\mathbf{Z}_i$ is the pair-level representation of pair $(X^{(p)}, X^{(h)})_{i}$ from Cross Attention module. Supervised contrastive loss $\mathcal{L}_{SCL(pair)}$ is calculated for every sentence pair among the batch $\mathcal{I}$. To minimize contrastive loss $\mathcal{L}_{SCL(pair)} $, the similarity of pairs in the same class should be as large as possible, and the similarity of negative examples should be as small as possible.

\subsubsection{\textbf{Cross-entropy Loss}}
Supervised contrastive loss mainly focuses on separating each pair apart from the others of different classes, whereas there is no explicit force in discriminating contradiction, neutral, and entailment. Therefore, we adopt the softmax-based cross-entropy to form the classification objective:
\begin{equation}
  \mathcal{L}_{CE} = CrossEntropy(\mathbf{W}\mathbf{Z} + \mathbf{b}, y),
\end{equation}
where $\mathbf{W}$ and $\mathbf{b}$ are trainable parameters. $\mathbf{Z}$ is the pair-level representation from Cross Attention module and $y$ is the corresponding label of the pair.

\subsubsection{\textbf{Overall Loss}} 
The overall loss is a weighted average of CE and the multi-level SCL loss, denoted as:
\begin{equation}
  \mathcal{L} = \mathcal{L}_{CE} + \alpha\mathcal{L}_{SCL(sent)} + \beta\mathcal{L}_{SCL(pair)},
\end{equation}
where $\alpha,\beta$ is a hyper-parameter to balance the objectives.

\section{EXPERIMENTAL SETUP}
\label{sec:results}
\subsection{Benchmark Dataset}
\label{sec:datasets}
We evaluate our model on three popular benchmarks: the Stanford Natural Language Inference (SNLI), the MultiGenre NLI Corpus (MultiNLI), and Sick. We also conduct cross-domain experiment while trained with all source domain data of one dataset and zero-shot transferred to the target domain of another dataset to evaluate the domain adaptation capability of the model. Detailed statistical information of these datasets is shown in Table \ref{table2}. Len(P) and Len(H) refer to the average length of \textbf{premise} and \textbf{hypothesis} respectively. MultiNLI(m) and MultiNLI(mm) indicate the matched and mismatched datasets respectively. We use classification accuracy as the evaluation metric.

\textbf{SNLI} The Stanford Natural Language Inference (SNLI) dataset contains 570k human-annotated sentence pairs, in which the premises are drawn from the captions of the Flickr30 corpus and hypotheses are manually annotated \cite{bowman-etal-2015-large}. This is the most widely used entailment dataset for NLI.

\textbf{MultiNLI} The corpus \cite{williams2018broad} is a new dataset for NLI, which contains 433k sentence pairs. Similar to SNLI, each pair is labeled with one of the following relationships: entailment, contradiction, or neutral. We use the matched dev set and mismatched dev set as our validation and test sets, respectively.

\textbf{Sick} This is a large data set on compositional meaning, annotated with subject ratings for both relatedness and entailment relation between sentences \cite{marelli2014sick}. The SICK data set consists of around 10000 English sentence pairs, each annotated for relatedness in meaning.

\textbf{SciTail} This is a textual entailment dataset derived from a science question answering (SciQ) dataset \cite{khot2018scitail}. The task involves assessing whether a given \textbf{premise} entails a given \textbf{hypothesis}. In contrast to other entailment datasets mentioned previously, the hypotheses in SciTail are created from science questions while the corresponding answer candidates and premises come from relevant web sentences retrieved from a large corpus. The dataset is only used for domain adaptation in this study. analyses.

\begin{table}[ht]
  \centering
  \caption{Statistics of Datasets}\label{table2}
\resizebox{0.9\linewidth}{!}{%
  \begin{tabular}{lccccc}
  \toprule
  \textbf{Dataset} & \textbf{Train} & \textbf{Dev}  & \textbf{Test} & \textbf{Len(P)} & \textbf{Len(H)} \\ 
  \midrule
  SNLI     & 549k  & 9.8k & 9.8k & 14     & 8         \\
  MultiNLI(m) & \multirow{2}{*}{392k}  & 9.8k & 9.8k & 22     & 11        \\
  MultiNLI(mm) &   & 9.8k & 9.8k & 22     & 11   \\ 
  Sick & 4.5k & 0.5k & 4.9k & 12 & 10 \\
  SciTail & 23.5k & 1.3k & 2.1k & 20 & 12 \\
  \bottomrule
  \end{tabular}%
  }
\end{table}

\subsection{Implementation Details}
We start from pre-trained checkpoints of BERT \cite{devlin2019bert} (uncased). We implement MultiSCL based on Huggingface's \texttt{transformers} package~\cite{wolf2019huggingface}. All experiments are conducted on 1 Nvidia GTX 3090 GPU. 

We train our models for 10 epochs with a batch size of 512 and temperature $\tau$ = 0.08 using an Adam optimizer \cite{kingma2014adam}. The hyper-parameter $\alpha$ and $\beta$ are set as 1 for combining objectives. The learning rate is set as 5e-5 for base models. Weight decay is used with a coefficient of 1e-5. The maximum sequence length is set to 128. All the experiments are conducted 5 times with different random seeds and we report the average scores. The hyperparameter $\eta$ is set to 0.1, which indicates the percent of the changed words in a sentence during the data augmentation module. We select Reordering and Dropout as augmentation strategies in the main experiments. 

\subsection{Baseline Models}
To analyze the effectiveness of our model, we evaluate several approaches for traditional NLI scenarios and state-of-the-art methods for low-resource NLI scenarios as baselines as follows on the above datasets.
\begin{enumerate}
    \item Traditional NLI baselines:
        \begin{itemize}
            \item InferSent \cite{conneau2017supervised} uses a BiLSTM network with max-pooling to learn generic sentence embeddings that perform well on several NLI tasks.
            \item ESIM \cite{chen-etal-2017-enhanced} is a previous state-of-the-art model for the natural language inference (NLI) task. It is a sequential model that incorporates the chain LSTM and the tree LSTM to infer local information between two sentences.
            \item BERT \cite{devlin2019bert} is the naturally bidirectional masked language model, configured with ‘bert-base-uncased’.
            \item PairSCL \cite{li2022pair} is a pair-level supervised contrastive learning approach with BERT as encoder. It adopts a Cross Attention module to learn the joint representations of the sentence pairs.
        \end{itemize}
    \item Low-resource NLI baselines:
        \begin{itemize}
            \item Gen-NLI \cite{ding2020discriminatively} is a generative classifier for NLI tasks. The model defines conditional probabilities assumed given premises and labels, parameterizing the distribution using a sequence-to-sequence model with attention and a replication mechanism. 
            \item MT-DNN \cite{liu2019multi} is a multi-task deep neural network for learning semantic representations across multiple natural language understanding tasks. 
        \end{itemize}
\end{enumerate}

\section{Experiment Results}
\label{sec:experiment}

\begin{table}[ht]
  \centering
  \caption{Performance on Test Datasets with Various Amounts of Training Data\tnote{*}}\label{table3}
  \resizebox{0.96\linewidth}{!}{%
  \begin{tabular}{l|lcccccc}
  \toprule
  Dataset & Model & 5 & 20 & 100 & 500 & 1000 & all \\ \hline
  \multirow{7}{*}{SNLI} & InferSent  & 37.5 & 39.6 & 44.1 & 56.0 & 63.9 & 84.5 \\ 
   & ESIM  & 38.4 & 38.6 & 46.7 & 58.2 & 65.4 & 87.6 \\
   & BERT  & 33.4 & 37.3 & 47.4 & 70.1 & 78.7 & 90.6 \\
   & PairSCL & 35.7 & 38.4 & 48.3 & 71.0 & 79.4 & 91.9 \\ \cline{2-8}
   & GenNLI & 43.5 & 45.6 & 50.6 & 60.6 & 64.2 & 82.2 \\
   & MT-DNN & 40.2 & 41.4 & 47.5 & 72.4 & 77.3 & 91.0 \\ 
   & MultiSCL & \textbf{45.3} & \textbf{48.2} & \textbf{55.4} & \textbf{73.7} & \textbf{80.1} & \textbf{92.2}\\ \hline
  \multirow{7}{*}{MNLI} & InferSent  & 34.1 & 33.7 & 35.2 & 44.9 & 47.9 & 70.4 \\ 
   & ESIM  & 36.9 & 35.4 & 40.5 & 49.8 & 54.2 & 76.7 \\
   & BERT  & 33.0 & 34.9 & 41.6 & 63.6 & 68.5 & 83.3 \\
   & PairSCL & 32.1 & 34.4 & 40.8 & 64.1 & 68.9 & 84.6 \\ \cline{2-8}
   & GenNLI & 44.1 & 47.1 & 49.0 & 60.6 & 63.4 & 67.5 \\
   & MT-DNN & 39.8 & 42.5 & 45.6 & 65.2 & 69.1 & 81.4 \\ 
   & MultiSCL & \textbf{47.2} & \textbf{49.3} & \textbf{52.7} & \textbf{66.4} & \textbf{70.6} & \textbf{85.4}\\ \hline
  \multirow{7}{*}{SICK} & InferSent  & 35.5 & 46.3 & 60.2 & 73.2 & - & 83.6 \\ 
   & ESIM  & 34.5 & 48.4 & 62.9 & 75.4 & - & 84.6 \\
   & BERT  & 36.7 & 56.7 & 63.6 & 78.6 & - & 86.0 \\
   & PairSCL & 35.8 & 55.2 & 63.1 & 79.2 & - & 86.5 \\ \cline{2-8}
   & GenNLI & 50.6 & 64.7 & 68.7 & 75.2 & - & 80.4 \\
   & MT-DNN & 46.4 & 61.9 & 64.2 & 78.3 & - & 85.8 \\ 
   & MultiSCL & \textbf{54.7} & \textbf{67.5} & \textbf{71.7} & \textbf{81.4} & - & \textbf{87.1} \\ \bottomrule
  \end{tabular}
  }
\end{table}

\subsection{Main Results}
We first empirically compare MultiSCL with baselines for three NLI datasets in low-resource scenarios. We construct smaller training sets by randomly selecting 5, 20, 100, 500, and 1000 instances per class, and then train separate models across these different-sized training sets. Table \ref{table3} shows the average performance
and standard deviation of the three runs of our model in comparison with the baselines on SNLI, MNLI, and SICK.\footnote{SICK does not have results in the 1000 column because the 'contradiction' label has only 665 instances.} The best result for each dataset and data amount is shown in bold. We also conduct the students paired t-test and the p-value of the significance test between the results of MultiSCL and GenNLI is less than 0.05 and 0.01, respectively. The results show that MultiSCL outperforms baselines on all data volume settings for all three datasets. When using training sets with 5/20/100 instances per class on three datasets, MultiSCL outperforms the state-of-the-art model by 3\%, 2.5\%, and 3.8\% respectively, which proves that MultiSCL can better capture the latent semantic representations by multi-level contrastive learning in low-resource scenarios. We can observe that transformer-based models such as BERT and PairSCL have poor performance results when the training data is less than 100 instances, even lower than LSTM-based models such as ESIM and InferSent. This shows that large-scale pre-trained models require a large amount of supervised data to be finetuned.

When the training set gets larger, the performance gap between MultiSCL and baselines does shrink. When trained with 500/1000/all instances per label, the accuracy exceeds the state-of-the-art model by 1.6\%, 1.1\%, and 0.6\%, which shows that MultiSCL has a more significant advantage compared with other models when trained with a smaller amount of training data. Furthermore, MultiSCL outperforms our previous work PairSCL by 10.9\% on average with trained less than 100 instances per class. The performance gains are due to the data augmentation module and the stronger ability to learn sentence-level latent embeddings in semantic space. The encoder of MultiSCL can capture sentence-level semantics effectively by the specifically-designed contrastive signal -- regarding the entailment pairs as positives and the contradiction pairs as negatives with limited training data. When trained with full training data, MultiSCL exceeds PairSCL by 0.6\%. We will further analyze the role of each module of MultiSCL in more detail in Section \ref{ablation}.

\subsection{Ablation Study}
\label{ablation}
We run extensive ablations to better understand the contribution of each key component of MultiSCL. We conduct experiments with the training set of 5/20/100/500 instances per class on SNLI and select Reordering and Dropout as augmentation strategies. The results are shown in Table \ref{table5}.

\begin{table}[!ht]
  \centering
  \caption{Ablation Study on SNLI}\label{table5}
  \resizebox{0.95\linewidth}{!}{%
  \begin{tabular}{lcccc}
  \toprule
  \textbf{Model}              & 5 & 20 & 100 & 500 \\ \hline
  MultiSCL (-Data Augmentation) & 39.5 & 44.8 & 52.1 & 71.5 \\
  MultiSCL (-Cross Attention) & 42.3 & 46.0 & 52.8 & 69.5   \\
  MultiSCL (-SCL(sent) loss)     & 41.0 & 45.1 & 51.7 & 71.2  \\
  MultiSCL (-SCL(pair) loss) & 42.7 & 44.3 & 52.5 & 70.6 \\
  MultiSCL (-CE loss) & 44.2 & 47.0 & 54.8 & 73.7  \\
  MultiSCL      & \textbf{45.3} & \textbf{48.2} & \textbf{55.4} & \textbf{73.7}  \\ \bottomrule
  \end{tabular}
  }
\end{table}

After removing Data Augmentation module, the performance of MultiSCL is reduced by an average of 2.94\%. Moreover, we can observe that the importance of Data Augmentation module gradually decreases as the number of training instances increases. When training with 5 instances per class, the accuracy decreases by 5.8\% on the test set. This indicates that when the training data is small, augmenting the sentences can bring larger enhancements to the model. The reason is that Data Augmentation module can create challenging views which allow the encoder to learn the semantic representation better with multi-level contrastive learning. After removing Cross Attention mechanism, the model simply concatenates the representation of two sentences to obtain the representation of the sentence pair. The performance decreases by 3.0\%, 2.2\%, 2.6\%, and 3.2\% respectively with different sizes of training data, which shows the joint representation obtained by cross attention can well characterize the relationship between the sentence pair. Without the sentence-level supervised contrastive learning loss, the accuracy of our model is decreased by 4.3\%, 3.1\%, 3.7\%, and 2.5\% with 5/20/100/500 training data per class. This demonstrates that by regarding entailment sentence pairs as positive samples and contradiction pairs as negative samples allows the encoder to discriminate the semantic difference between sentences. After removing pair-level supervised contrastive learning loss, the performance decreases by 2.5\% on average. The reason is that the contrastive learning objective can learn the discrepancy between the sentence pairs of different classes by pulling the sentence pairs from the same class together and pushing the pairs of different classes further apart. The test accuracy decreases by 0.7\% on average without the cross-entropy loss.

\begin{figure}  
\centering
		\begin{minipage}[t]{0.47\linewidth}
			\centering
			\includegraphics[width=1\linewidth]{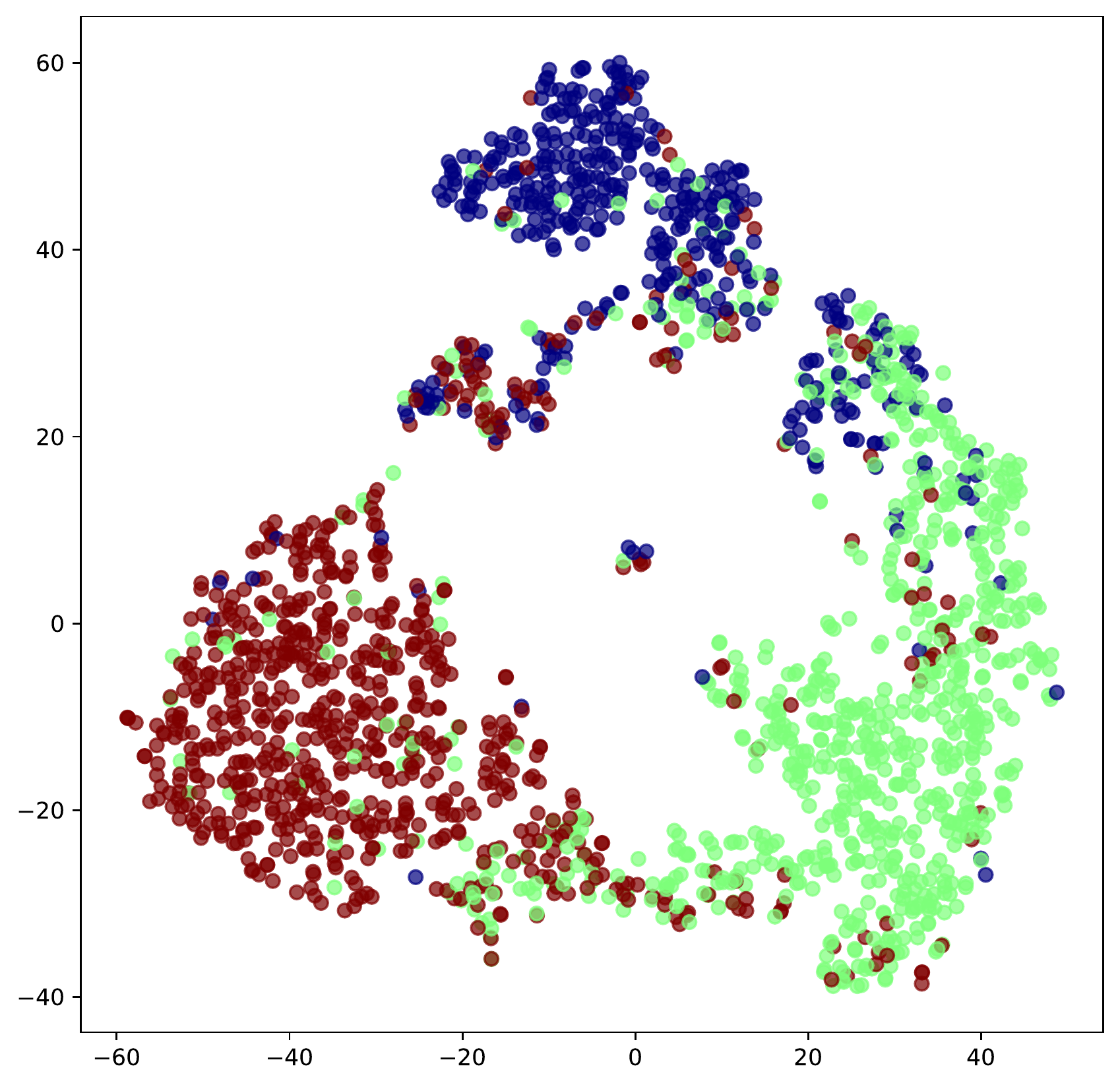}
		\end{minipage}
		\begin{minipage}[t]{0.48\linewidth}
			\centering
			\includegraphics[width=1\linewidth]{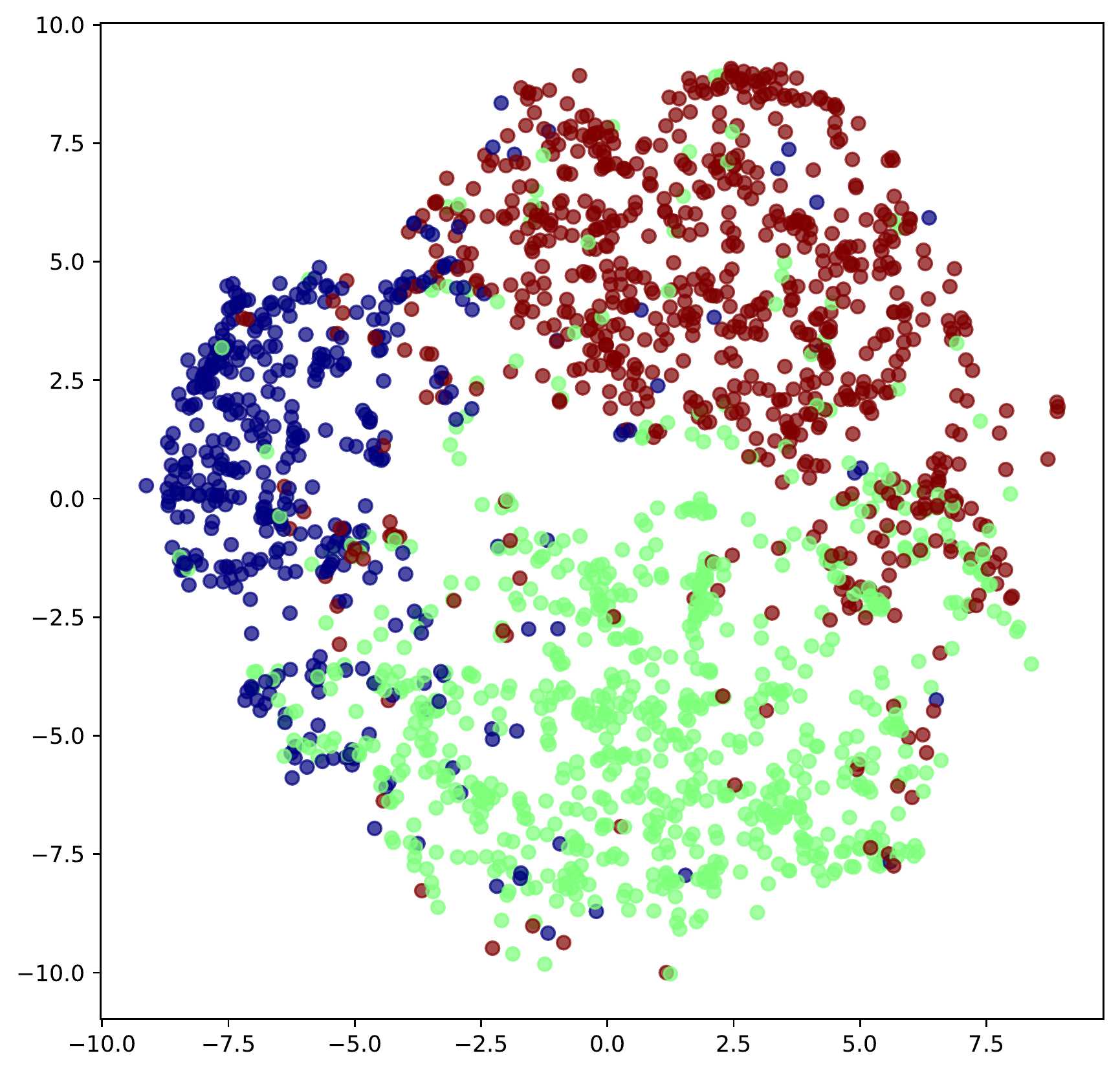}
		\end{minipage}
    \caption{\label{tsne} Visualization of the SNLI representations using t-SNE for MultiSCL (left) and MultiSCL w/o Cross Attention Module (right).}
\end{figure}

\section{Qualitative Analysis}
In this section, we further conduct extensive experiments to understand the inner workings of MultiSCL.
\subsection{Analysis of Cross Attention Module} To further investigate the effect of Cross Attention module, we conduct the t-SNE visualization experiments of the representations $\mathbf{Z}$ for the sentence pairs on SNLI test set. Figure \ref{tsne} illustrates that the contrastive loss can map the pairs of the same category closer together in the embedding space while pushing apart negative pairs in different classes. The right part of Figure \ref{tsne} shows the results after removing Cross Attention module. It is clear that the representations of the same category with Cross Attention module in left part of Figure \ref{tsne} are better grouped together compared to the representations in right part of Figure \ref{tsne}. That indicates Cross Attention module can learn the joint representation between \textbf{premise} and \textbf{hypothesis} very well. The representations of the positive pairs obtained from Cross Attention module can be mapped together in the semantic space by contrastive learning. This could well explain why removing Cross Attention module would give a high accuracy drop.

\begin{figure}[ht]
  \centering
  \includegraphics[width=1.0\linewidth]{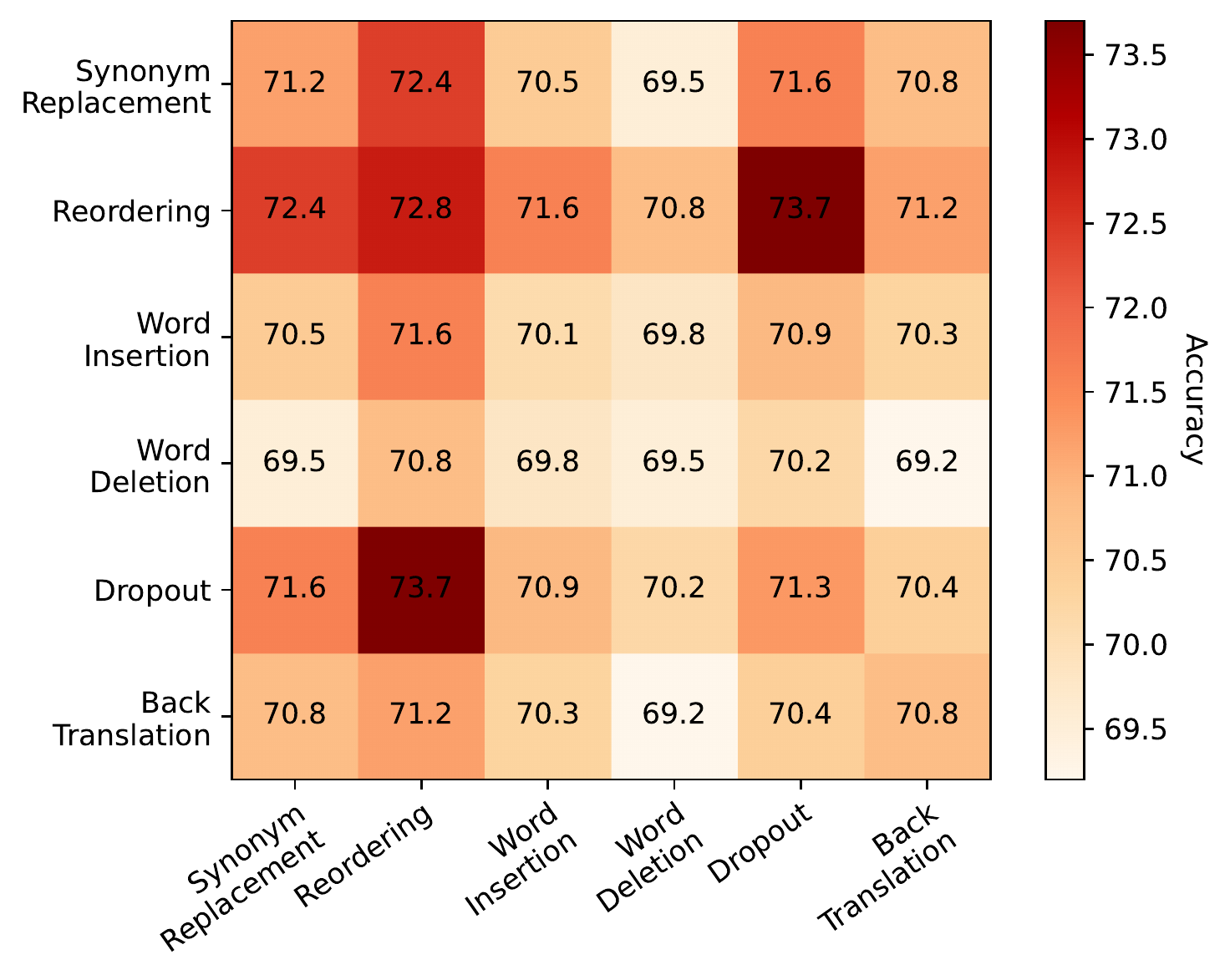} 
  \caption{\label{da} The performance visualization with different combinations of data augmentation strategies. The row indicates the 1st data augmentation strategy while the column indicates the 2nd data augmentation strategy.}
\end{figure}

\subsection{Analysis of Data Augmentation Module}
In this section, we first analyze the effect of different combinations of data augmentation strategies on SNLI with 500 training instances per class. As described in Section \ref{data augmentation}, we consider six options for each augmentation, including Synonym Replacement, Reordering, Word Insertion, Word Deletion, Dropout, and Back Translation, resulting in $6\times6$ combinations.

The results are shown in Figure \ref{da}. When using Word Deletion, MultiSCL has poorer performance than other strategies on average. The performance decreases to the lowest with an accuracy of 69.2\% with Word Deletion and Back Translation. We argue that Word Deletion and Back Translation may change the meaning compared with the original sentences, resulting in an unexpected change in the relationship between \textbf{premise} and \textbf{hypothesis}. In this situation, MultiSCL may learn to misunderstand semantic representation with multi-level contrastive learning by incorrectly setting the positive and negative set. Compared to Word Deletion and Back Translation, Word Insertion and Synonym Replacement improves the performance but does not achieve the best result. These two strategies can create views that have the same meaning as the original sentence but don't introduce meaningful changes. Therefore, the model cannot construct effective positive/negative sets with these augmented views in contrastive learning.

We can observe that Reordering and Dropout are the two most effective strategies with an accuracy of 73.7\% (where Reordering is slightly better than Dropout). We argue that Reordering and Dropout can create challenging sentence pairs for contrastive learning without changing the semantic information. The augmented views are useful for contrastive learning without confusing the model, and thus improve the model robustness in low-resource scenarios. We adopt Reordering and Dropout as augmentation strategies in most experiments. 

\begin{table}[!ht]
  \centering
  \caption{Influence of $\eta$ in Data Agumentation Module}\label{table6}
  \begin{tabular}{cccccc}
  \toprule
  $\eta$           & 10\% & 20\% & 40\% & 60\% & 80\% \\ \hline
  MultiSCL      & \textbf{73.7} & 70.1 & 63.4 & 54.6 & 41.8  \\ \bottomrule
  \end{tabular}
\end{table}

Furthermore, we explore the effect of hyperparameter $\eta$ on SNLI with 500 training instances per class. $\eta$ is a hyperparameter that indicates the percent of the changed words in a sentence. The number of words changed $n$ is calculated with the formula $n$ = $\eta$l, where $l$ is the length of the sentence. The average length of \textbf{premise} and \textbf{hypothesis} is 14 and 8 for SNLI. If $\eta$ is set to 5\%, the number of changed words is less than 1. So we test the results when $\eta$ is set from 10\% to 80\%. Table \ref{table6} shows that the accuracy is highest when $\eta$ is set to 10\%. The accuracy decreases 3.6\% when $\eta$ is set to 20\%. As the value of $\eta$ becomes larger, the performance decreases dramatically. When the value of $\eta$ is 80\% in the extreme case, the accuracy of MultiSCL drops to a minimum of 41.8\%. The results are not surprising. When $\eta$’s value increases, the number of changed words increases which makes the augmented sentence and the original sentence more likely to have different meanings. In this situation, Data Augmentation module can introduce very serious noise to the model, so that the positive/negative pairs that we regard in multi-level contrastive learning do not work actually.

\begin{figure}[ht]
  \centering
  \includegraphics[width=0.98\linewidth]{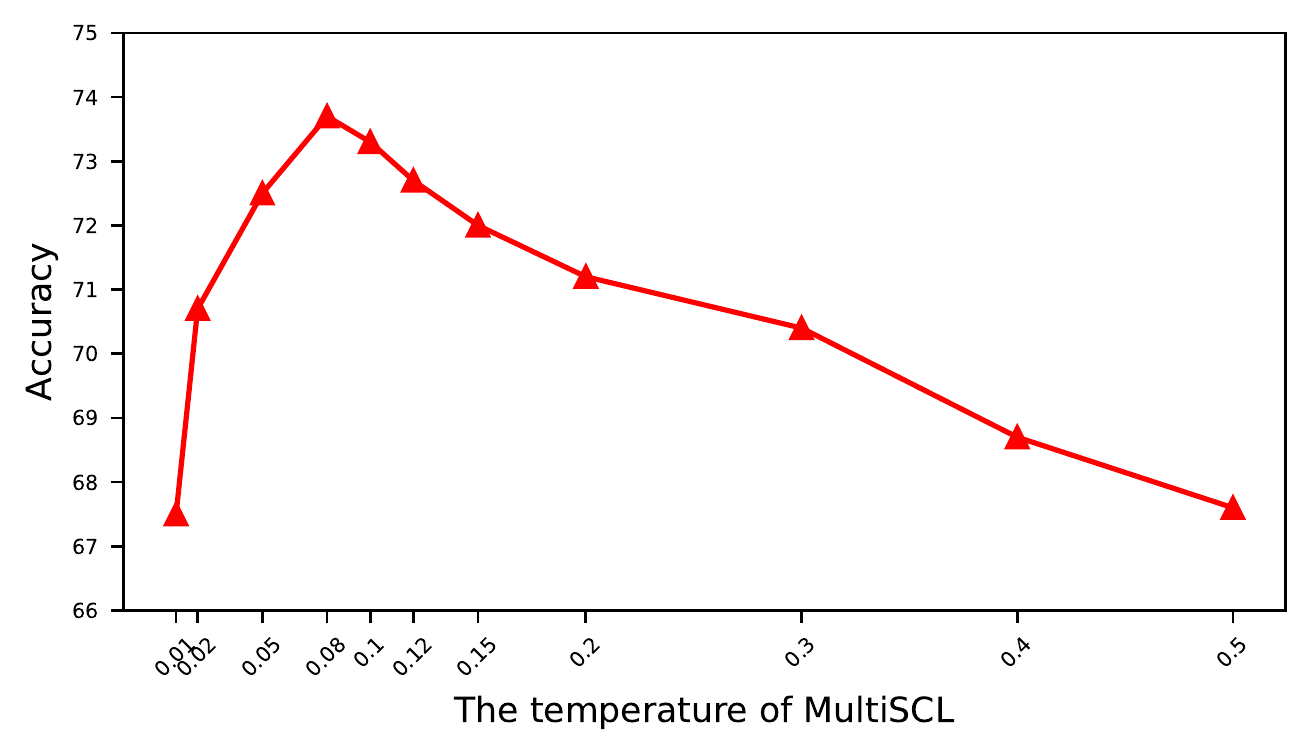} 
  \caption{\label{temp} The influence of different temperatures $\tau$ in MultiSCL. The best performance is achieved when the temperature is set to 0.08.}
\end{figure}

\subsection{Influence of Temperature} The temperature in sentence-level (Equation 11) and pair-level (Equation 12) contrastive loss is used to control the smoothness of the distribution normalized by softmax operation and thus influences the gradients when backpropagation. A higher temperature smooths the distribution while a low temperature scales up the dot-products and sharpens the distribution. In our experiments, we explore the influence of temperature $\tau$ on SNLI dataset with 500 training instances per class. The result is illustrated in Figure \ref{temp}.

As shown in the figure, we can observe that the performance of MultiSCL is very sensitive to the value of temperature $\tau$. As the temperature becomes higher, the performance of the model first improves and then decreases. Either too low or too high temperature will make our model perform badly. The optimal temperature value is 0.08 when MultiSCL has the highest accuracy of 73.7\%. This phenomenon again demonstrates that the temperature determines how much attention is paid to difficult negative samples in contrastive loss. The higher the temperature, the less attention is paid to difficult negative samples, while the lower the temperature, the model focuses more on negative samples that are very different from the anchor. We select 0.08 as the temperature in most of our experiments.

\begin{table}[htbp]
  \centering
  \caption{Influence of $\alpha$ and $\beta$ in Training Objective.}
    \resizebox{0.6\linewidth}{!}{%
    \begin{tabular}{ccccc}\toprule
    \diagbox{$\alpha$}{$\beta$} & 0.2   & 0.5   & \underline{1.0}     & 2.0 \\ \midrule
    0.2   & \cellcolor[rgb]{ 1,  .925,  .612}69.5 & \cellcolor[rgb]{ .98,  .773,  .529}70.2 & \cellcolor[rgb]{ .969,  .686,  .482}70.6 & \cellcolor[rgb]{ .937,  .467,  .365}71.6 \\
    0.5   & \cellcolor[rgb]{ .973,  .729,  .506}70.4 & \cellcolor[rgb]{ .945,  .529,  .4}71.3 & \cellcolor[rgb]{ .922,  .376,  .318}72 & \cellcolor[rgb]{ .918,  .353,  .306}72.1 \\
    \underline{1.0}     & \cellcolor[rgb]{ .941,  .51,  .388}71.4 & \cellcolor[rgb]{ .906,  .267,  .259}72.5 & \cellcolor[rgb]{ .867,  0,  .114}\textbf{73.7} & \cellcolor[rgb]{ .898,  .2,  .224}72.8 \\
    2.0    & \cellcolor[rgb]{ .949,  .553,  .412}71.2 & \cellcolor[rgb]{ .902,  .224,  .235}72.7 & \cellcolor[rgb]{ .89,  .157,  .2}73 & \cellcolor[rgb]{ .937,  .486,  .376}71.5 \\ \bottomrule
    \end{tabular}%
    }
  \label{tabel7}%
\end{table}%

\subsection{Influence of Hyper-parameters in Training Objective}
To investigate the effect of hyper-parameters $\alpha$ and $\beta$ in Equation 15, we conduct the experiments with 500 instances per class on SNLI by setting different values. The results are illustrated in Table \ref{tabel7} and we underline the setting used for all our experiments. The hyper-parameters $\alpha$ and $\beta$ are used to balance sentence-level contrastive loss and pair-level contrastive loss in the training objective. MultiSCL focuses more on semantic similarity discrimination of sentences with a larger value of $\alpha$, while the model is paying more attention to learning the discrepancy of sentence pairs in different categories with a larger value of $\beta$.

Table \ref{tabel7} indicates a trade-off between the sentence-level semantic encoding capability and the pair-level reasoning capability of MultiSCL. When the values of both $\alpha$ and $\beta$ are 0.2, MultiSCL almost removes contrastive learning and uses only cross-entropy loss, and the accuracy decreases by 4.2\%. The performance of MultiSCL keeps improving as the value of $\alpha$ and $\beta$ increases until the highest accuracy of 73.7\% with $\alpha=\beta=1.0$. However, as the values of $\alpha$ and $\beta$ continue to increase, the accuracy begins to decrease. This result is not surprising, especially considering that the joint representation of a contradiction pair in pair-level contrastive learning is obtained from two sentences that are regarded as negative sets in the sentence-level contrastive learning. Focusing too much on the pair-level classification objective, i.e., using larger $\beta$ values, can hurt the embeddings of the sentence from the encoder. On the other hand, focusing overly on separating semantically dissimilar sentences also affects the discrimination of sentence pairs in different categories. We set $\alpha$ and $\beta$ as 1.0 for all our experiments with effective multi-level supervised contrastive learning.

\subsection{Case Study}
\begin{table*}[!ht]
  \centering
  \caption{Selected Examples from SNLI}\label{table8}
  \begin{tabular}{cllccc}
  \toprule
  ID           & \textbf{Premise} & \textbf{Hypothesis} & GenNLI & MultiSCL & Gold \\ \hline
  A      & Two kids are standing in the ocean hugging each other. & Two kids enjoy their day at the beach. & E & N & N  \\ 
  B & They are sitting on the edge of a fountain. & The fountain is splashing the persons seated. & E & N & N \\ \midrule
  C & An old man with a package poses in front of an advertisement. & A man poses in front of an ad. & N & E & E \\
  D & A young family enjoys feeling ocean waves lap at their feet. & A family is at the beach. &N & E & E \\ \midrule
  E & A man and woman sit at a cluttered table. & The table is neat and clean.& E & C & C \\
  F & A race car sits in the pits. & The car is going fast.& N & C & C \\ \bottomrule
  \end{tabular}
\end{table*}
To illustrate the advantages of our model in more detail, we conduct a case study. Table \ref{table8} includes some examples from the SNLI test set, where MultiSCL successfully predicts the relation and GenNLI fails. In examples A and B, both sentences contain phrases that are either identical or highly lexically related (e.g. “Two kids”, “ocean/beach”, and "fountain"), which confuses GenNLI's judgment as entailment category. MultiSCL can infer the correct relationship of Neutral by capturing the difference between words in \textbf{premise} and \textbf{hypothesis} with Cross Attention module. For examples C and D, GenNLI regarded their relationships as Neutral but the gold labels are Entailment. The reason may be that the two sentences do not have some identical words, so GenNLI cannot easily recognize their semantic similarity. However, MultiSCL can correctly capture the semantics of the sentences through multi-level contrastive learning. For example E, GenNLI predicts the relationship as Entailment while MultiSCL can infer the correct relationship as Contradiction from "cluttered" and "neat and clean". For example F, MultiSCL can predict the relationship as Contradiction from "sits" and "going fast" while GenNLI considers their relationship to be Neutral. These results show that MultiSCL can understand the semantic information by capturing the interaction of words in \textbf{premise} and \textbf{hypothesis}. Furthermore, MultiSCL can better infer the relationship of the challenging pairs with the multi-level contrastive learning.

\subsection{Error Analysis}
To analyze the limitation of MultiSCL, we select some very challenging cases in which both GenNLI and MultiSCL cannot infer the relationships correctly in low-resourse scenarios. For example, MultiSCL predicts the relationship as entailment for the contradiction pair (\textbf{premise}: A person wearing a straw hat, standing outside working a steel apparatus. \textbf{hypothesis}: A person is burning a straw hat.). The most likely reason is that all the words in \textbf{hypothesis} except "burning" are included in \textbf{premise}. In this situation, MultiSCL ignores the different words ("wearing" and "burning") and simply assumes that \textbf{premise} and \textbf{hypothesis} are nearly identical. Therefore, the model predicts the relationship as entailment.
Another example of model misclassification is the pair (\textbf{premise}: Two women having drinks and smoking cigarettes at the bar.  \textbf{hypothesis}: Three women are at a bar.). The gold label is contradiction but MultiSCL believes their relationship is entailment. We can observe that the two sentences describe basically the same scenario, only the number of people mentioned is different. This indicates that our model ignores differences in count words when the semantics of the sentences are similar.

\begin{table}[htbp]
  \centering
  \caption{Domain Adaptation Performance}
     \resizebox{0.9\linewidth}{!}{%
    \begin{tabular}{cclc}
    \toprule
    Source Dataset & Target Dataset & Model & Accuracy \\
    \midrule
    \multirow{4}[2]{*}{SNLI} & \multirow{4}[2]{*}{MNLI} & BERT  & 67.0 \\
          &       & GenNLI &  61.4\\
          &       & MT-DNN &  69.3\\
          &       & Multi-SCL & \textbf{77.5} \\
    \midrule
    \multirow{4}[2]{*}{SNLI} & \multirow{4}[2]{*}{SciTail} & BERT  & 49.0 \\
          &       & GenNLI &  42.7\\
          &       & MT-DNN &  49.3\\
          &       & Multi-SCL &  \textbf{60.2}\\
    \midrule
    \multirow{4}[2]{*}{MNLI} & \multirow{4}[2]{*}{SNLI} & BERT & 77.4 \\
          &       & GenNLI &  71.8\\
          &       & MT-DNN &  78.5\\
          &       & Multi-SCL &  \textbf{85.0}\\
    \midrule
    \multirow{4}[2]{*}{MNLI} & \multirow{4}[2]{*}{SciTail} & BERT  & 49.8 \\
          &       & GenNLI &  47.6\\
          &       & MT-DNN &  50.3\\
          &       & Multi-SCL &  \textbf{58.2}\\
    \midrule
    \multirow{4}[2]{*}{SciTail} & \multirow{4}[2]{*}{SNLI} & BERT  &  39.1\\
          &       & GenNLI &  35.9\\
          &       & MT-DNN &  40.1\\
          &       & Multi-SCL &  \textbf{49.2}\\
    \midrule
    \multirow{4}[2]{*}{SciTail} & \multirow{4}[2]{*}{MNLI} & BERT  &  41.0\\
          &       & GenNLI &  37.8\\
          &       & MT-DNN &  42.5\\
          &       & Multi-SCL &  \textbf{51.4}\\
    \bottomrule
    \end{tabular}%
    }
  \label{table4}%
\end{table}%

\section{Domain Adaptation Results}
To investigate the performance of the model in low-resource scenarios more deeply, we conduct the domain adaptation experiments. We evaluate MultiSCL with BERT, GenNLI, and MT-DNN by domain adaptation between SNLI, MNLI, and Scitail. The models are trained to convergence using all the source domain data and zero-shot transferred to the target domain, which can evaluate the ability for domain-independent reasoning. The domain adaptation results are illustrated in Table \ref{table4}. MultiSCL outperforms all baselines by 8.5\% on average for all six domain adaptation settings. Furthermore, we can find that all models perform the best for domain adaptation of SNLI and MNLI in various combinations of source and target datasets. The reason is that MNLI is modeled on SNLI but differs in that it covers a range of genres including transcribed speech, fiction, and government reports. Therefore, the domain similarity between SNLI and MNLI is higher and the models are more likely to transfer semantic knowledge. Moreover, the model is easier to transfer from MNLI to SNLI, compared with SNLI to MNLI.

For the domain adaptation results between SNLI/MNLI and SciTail, the accuracy is lower compared to other cases. The reason is that SciTail is a textual entailment dataset from the science domain and domain information differs significantly from other datasets. In addition, SciTail is a smaller dataset that only contains 23.5k training data as introduced in \ref{sec:datasets}. Therefore, it is a challenge to transfer between dissimilar domains such as SciTail to SNLI/MNLI. Our model, Multi-SCL, outperforms the state-of-the-art model by 8.8\% on average in six domain adaptation scenarios. We can observe that MulitSCL achieves an outstanding performance of 49.2\% and 51.4\% adapted from SciTail to SNLI/MNL, exceeding the state-of-the-art model by 10.1\% and 10.4\%, respectively. These results indicate that MultiSCL has a stronger ability to learn domain invariant latent representations through multi-level contrastive learning. MultiSCL can accurately characterize the sentence pairs in the semantic space by the specifically-designed contrastive signal and zero-shot transferred to different domains.

\section{Conclusion and Future Work}
\label{sec:conclusions}
In this paper, we propose a multi-level supervised contrastive learning approach (MultiSCL) for low-resource NLI. We adopt a data augmentation module to create different views for input sentences. MultiSCL leverages sentence-level contrastive learning to enhance the capability of semantic modeling by naturally taking the entailment \textbf{hypothesis} as the corresponding \textbf{premise}'s positive set and the contradiction \textbf{hypothesis} as the negative set. The cross attention module is designed to learn the joint representations of the sentence pairs. The pair-level contrastive learning objective is aimed to distinguish the varied classes of sentence pairs by pulling those in one class together and pushing apart the pairs in other classes. We evaluate MultiSCL on three popular NLI datasets in low-resource settings. The experiment results show that MultiSCL outperforms the previous state-of-the-art method performance by 3.1\%. For the domain adaptation tasks, the accuracy of MultiSCL exceeds existing models by 8.5\% on average. We carefully study the components of MutliSCL and show the effects of different parts. We also compare multiple combinations of data augmentation strategies and provide fine-grained analysis of several hyper-parameters to interpret how our approach works.

In future work, we intend to exploit using contrastive learning to obtain representations that can more accurately express the relationship between sentences in low-resource settings. Furthermore, we will investigate more effective data augmentation methods for texts. Other future work will be to measure the performance of MultiSCL on adversarial and similarly challenging NLI datasets. We hope our work will provide a new perspective for future research on contrastive learning.


%
\bibliographystyle{IEEEtran} 
\bibliography{IEEEabrv,refs}

\begin{IEEEbiography}[{\includegraphics [width=1in,height=1.25in,clip, keepaspectratio]{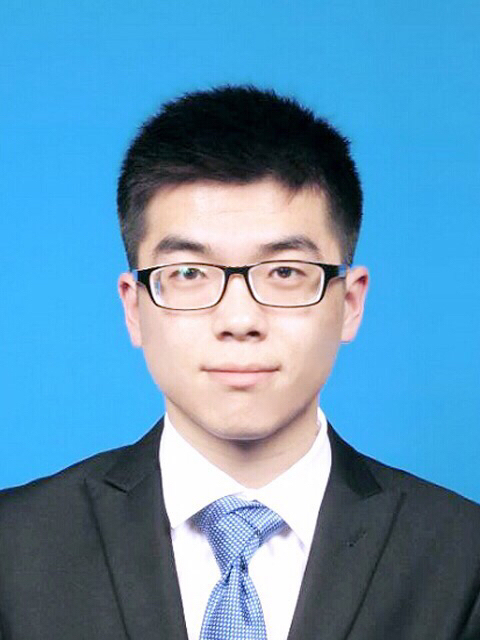}}] {Shu'ang Li} received the B.S. degree from Tsinghua University, Beijing, China, in 2018. He is currently working towards the Ph.D. degree with the School of Software, Tsinghua University, Beijing, China. His research interests include text classification and deep learning. \end{IEEEbiography}
\begin{IEEEbiography}[{\includegraphics [width=1in,height=1.25in,clip, keepaspectratio]{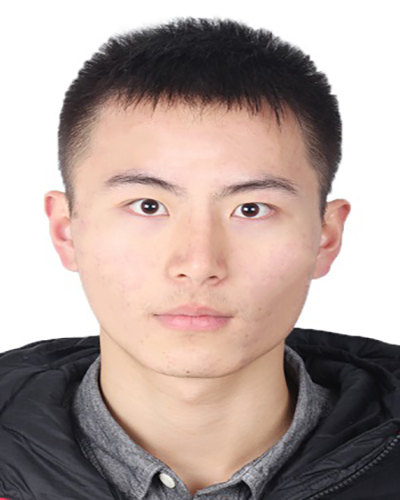}}] {Xuming Hu} received the BE degree in Computer Science and Technology, Dalian University of Technology. He is working towards the PhD degree at Tsinghua University. His research interests include natural language processing and information extraction. \end{IEEEbiography}
\begin{IEEEbiography}[{\includegraphics [width=1in,height=1.25in,clip, keepaspectratio]{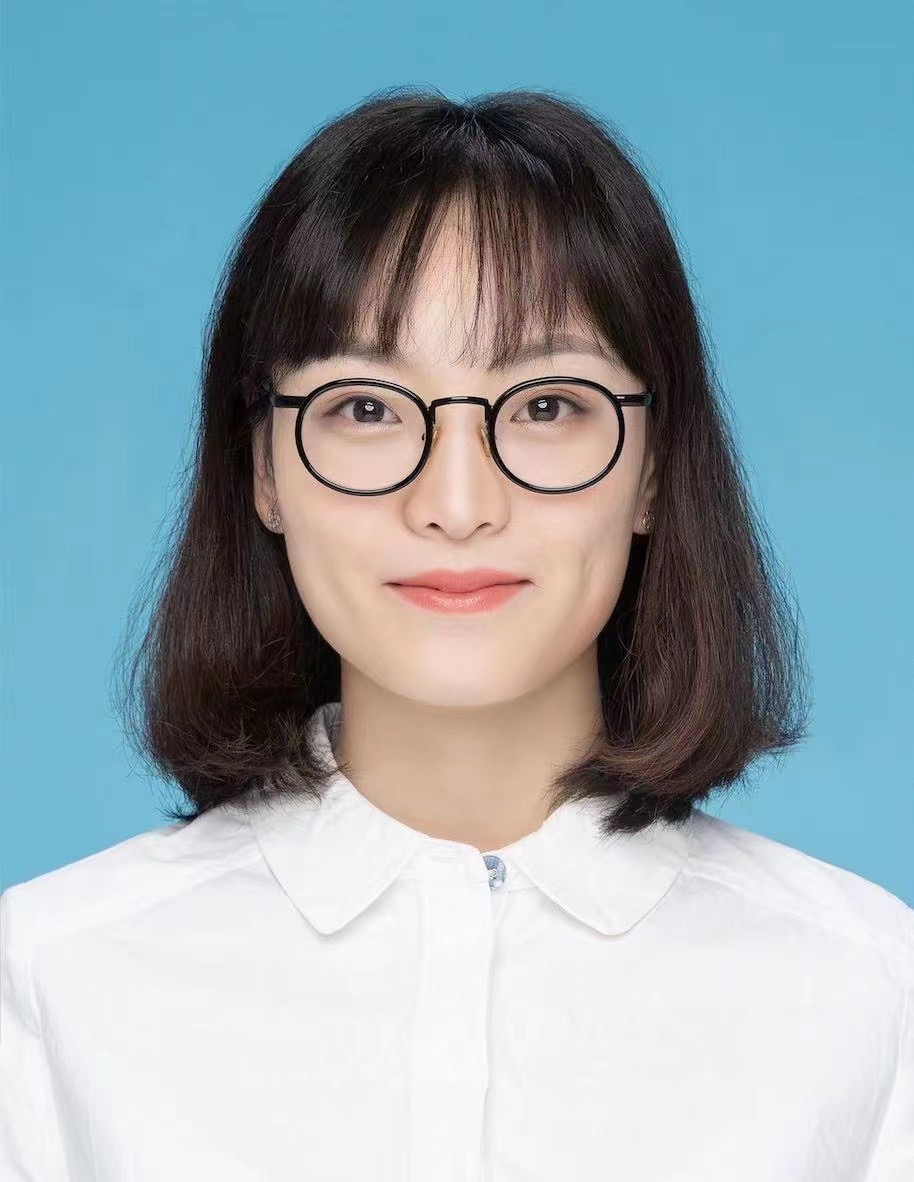}}] {Li Lin} received the B.S. degree from Xi'an Jiaotong University, Shaanxi, China, in 2016. She received the Ph.D. degree with the School of Software, Tsinghua University, Beijing, China, in 2022. Her research interests include sequence modeling and deep learning. \end{IEEEbiography}
\begin{IEEEbiography}[{\includegraphics [width=1in,height=1.25in,clip, keepaspectratio]{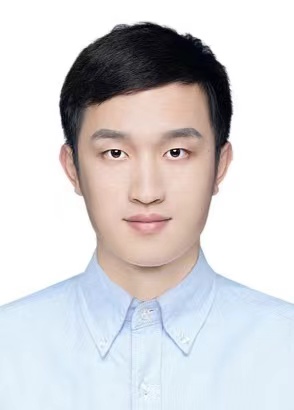}}] {Aiwei Liu} received the BE degree in Software Engineering, Nanjing University. He is working toward the PhD degree at Tsinghua University. His research interests include natural language processing and semantic parsing. \end{IEEEbiography}
\begin{IEEEbiography}[{\includegraphics [width=1in,height=1.25in,clip, keepaspectratio]{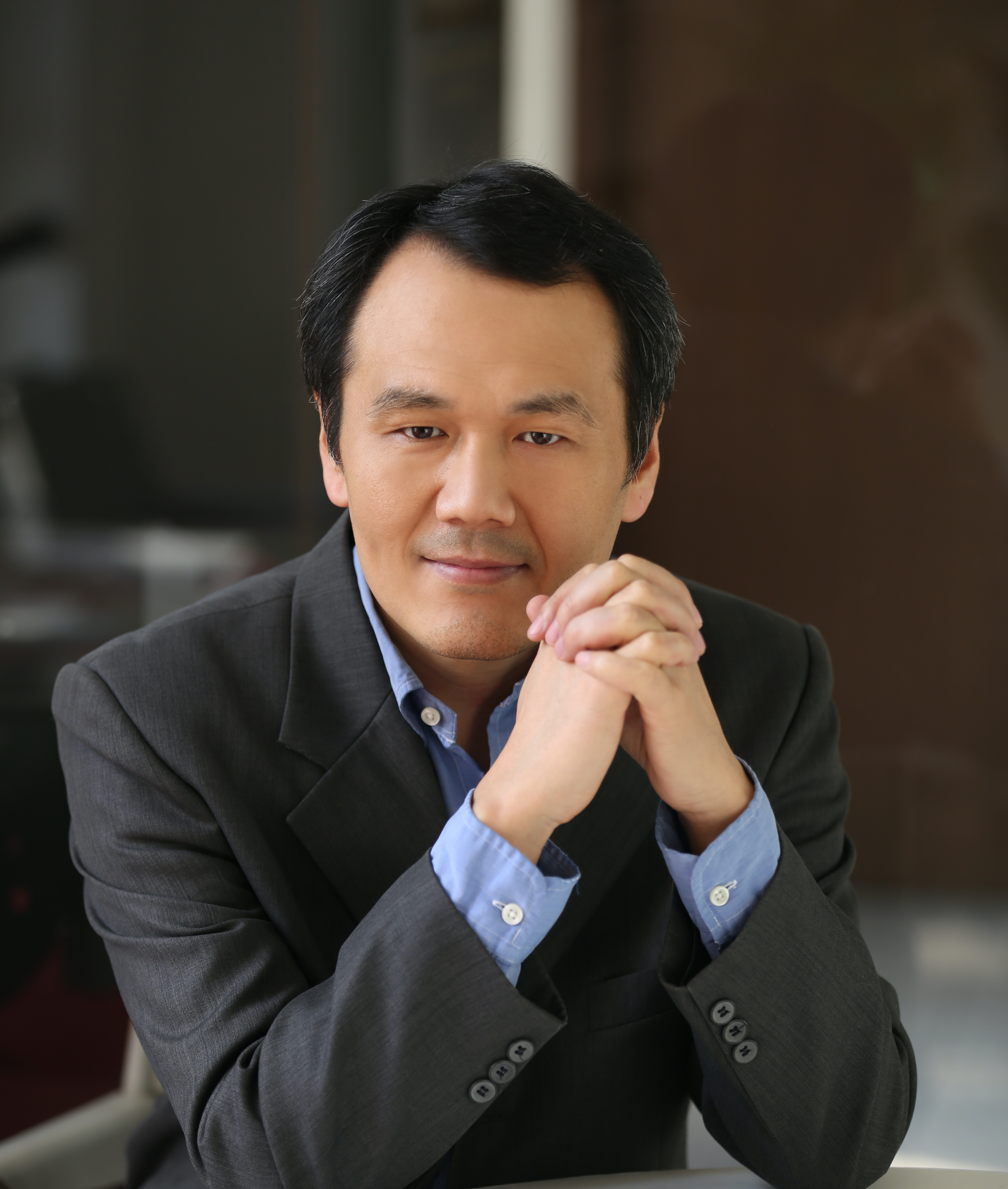}}] {Lijie Wen} received the B.S. degree, the M.S. degree, and the Ph.D. degree in Department of Computer Science and Technology, Tsinghua University, Beijing, China, in 2000, 2003, and 2007 respectively. He is currently an associate professor at School of Software, Tsinghua University. His research interests are focused on process data management, lifecycle management of computational workflow, and natural language processing. He has published more than 150 academic papers on conferences and journals, which are cited more than 4100 times by Google Scholar. \end{IEEEbiography}
\begin{IEEEbiography}
[{\includegraphics [width=1in,height=1.25in,clip, keepaspectratio]{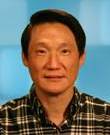}}] {Philip S. Yu} (Life Fellow, IEEE) is currently a Distinguished Professor and the Wexler Chair of information technology with the Department of Computer Science, University of Illinois Chicago (UIC), Chicago, IL, USA. Before joining UIC, he was with IBM Watson Research Center, where he built a world-renowned data mining and database department. He has authored or coauthored more than 780 papers in refereed journals and conferences. He holds or has applied for more than 250 U.S. Patents. His research interest include Big Data, including data mining, data stream, database, and privacy. He is a Fellow of ACM. Dr. Yu was the Editor-in-Chief of the ACM Transactions on Knowledge Discovery from Data during 2011–2017 and IEEE TRANSACTIONS ON KNOWLEDGE AND DATA ENGINEERING during 2001–2004. He was the recipient of several IBM honors including the two IBM Outstanding Innovation Awards, Outstanding Technical Achievement Award, two Research Division Awards, and 94th Plateau of Invention Achievement Awards.
\end{IEEEbiography}

\vfill
\end{document}